\documentclass[letterpaper]{article} % DO NOT CHANGE THIS
\usepackage{aaai2026}  % DO NOT CHANGE THIS
\usepackage{times}  % DO NOT CHANGE THIS
\usepackage{helvet}  % DO NOT CHANGE THIS
\usepackage{courier}  % DO NOT CHANGE THIS
\usepackage[hyphens]{url}  % DO NOT CHANGE THIS
\usepackage{graphicx} % DO NOT CHANGE THIS
\usepackage{natbib}  % DO NOT CHANGE THIS AND DO NOT ADD ANY OPTIONS TO IT
\usepackage{caption} % DO NOT CHANGE THIS AND DO NOT ADD ANY OPTIONS TO IT
\usepackage{algorithm}
\usepackage{algorithmic}

\usepackage{newfloat}
\usepackage{listings}
\DeclareCaptionStyle{ruled}{labelfont=normalfont,labelsep=colon,strut=off} % DO NOT CHANGE THIS
\floatstyle{ruled}
\newfloat{listing}{tb}{lst}{}
\floatname{listing}{Listing}
\usepackage{amsmath, amssymb}
\usepackage{makecell}
\usepackage{amsfonts}  % some math symbol  for example, /mathbb{}
\usepackage{graphicx}  % resize tables and figures
\usepackage{boldline}  % for table
\usepackage{multirow}  % for table
\usepackage{booktabs}  % for table
\usepackage{array}     % for table
\usepackage{xcolor}    % for table
\usepackage{arydshln}
\usepackage{makecell}
\usepackage{subcaption} % for figures

\title{Beyond Chains: Bridging Large Language Models and Knowledge Bases in Complex Question Answering}
\author{
    Yihua Zhu\textsuperscript{\rm 1,3},
    Qianying Liu\textsuperscript{\rm 3},
    Akiko Aizawa\textsuperscript{\rm 2,3},
    Hidetoshi Shimodaira\textsuperscript{\rm 1,4}
}
\affiliations{
    \textsuperscript{\rm 1}Kyoto University\\
    \textsuperscript{\rm 2}University of Tokyo\\
    \textsuperscript{\rm 3}NII LLMC\\
    \textsuperscript{\rm 4}RIKEN AIP\\
    zhu.yihua.22h@st.kyoto-u.ac.jp, ying@nii.ac.jp, aizawa@nii.ac.jp, shimo@i.kyoto-u.ac.jp
}

\usepackage{bibentry}
\begin{document}

\maketitle

\begin{abstract}
Knowledge Base Question Answering (KBQA) aims to answer natural language questions using structured knowledge from KBs. While LLM-only approaches offer generalization, they suffer from outdated knowledge, hallucinations, and lack of transparency. Chain-based KG-RAG methods address these issues by incorporating external KBs, but are limited to simple chain-structured questions due to the absence of planning and logical structuring.
Inspired by semantic parsing methods, we propose \textbf{PDRR}: a four-stage framework consisting of \textbf{P}redict, \textbf{D}ecompose, \textbf{R}etrieve, and \textbf{R}eason. Our method first \textbf{predicts} the question type and \textbf{decomposes} the question into structured triples. Then \textbf{retrieves} relevant information from KBs and guides the LLM as an agent to \textbf{reason} over and complete the decomposed triples.
Experimental results show that our proposed KBQA model, PDRR, consistently outperforms existing methods across different LLM backbones and achieves superior performance on various types of questions.
\end{abstract}

% Uncomment the following to link to your code, datasets, an extended version or similar.
% You must keep this block between (not within) the abstract and the main body of the paper.
\begin{links}
    \link{Code}{https://github.com/YihuaZhu111/PDRR}
    \link{Extended version}{https://arxiv.org/abs/2505.14099}
\end{links}

\section{Introduction}

%.....figure drawbacks of TOG

\begin{figure}[t]
\centering

\includegraphics[width=1\columnwidth]{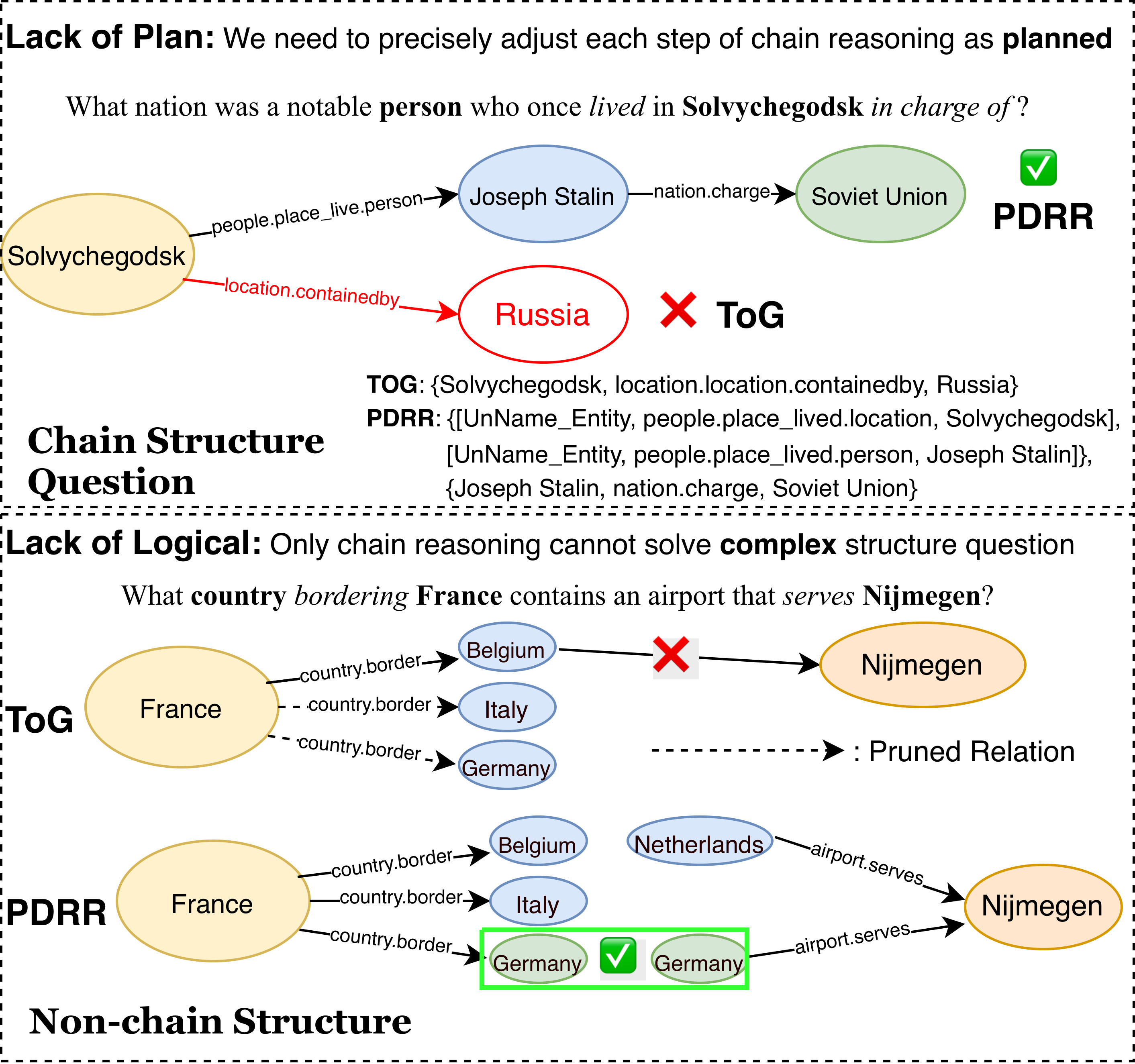}
\caption{Drawbacks of ToG (a chain-based KG-RAG approach).
ToG and similar chain-based KG-RAG methods lack a planning module for explicit reasoning control and are limited to chain-type questions due to their insufficient logical structuring. Our PDRR framework resolves both issues.}
\label{fig:drawbacks_TOG}
\end{figure}

%........Figure

Knowledge bases (KBs) offer rich, structured repositories of world knowledge, where facts are organized as (subject, relation, object) triples. Large-scale KBs such as Freebase \cite{bollacker2008freebase}, DBpedia \cite{lehmann2015dbpedia}, Wikidata \cite{pellissier2016freebase}, and YAGO \cite{suchanek2007yago} provide useful resources for downstream applications. Knowledge base question answering (KBQA) leverages this structured information to translate natural language question queries into precise, verifiable answers, which poses challenges since it requires accurate multi-hop reasoning. This capability is essential for domains that require accurate and verifiable retrieval of facts.

In response to the costly human effort required to annotate training data, recent studies have turned to leveraging large language models (LLMs) based methods for KBQA, for example, IO \cite{brown2020language}, CoT \cite{wei2022chain}, and SC \cite{wang2022self}. These methods harness broad, pre-trained knowledge of LLMs and exhibit strong generalization across datasets. Nonetheless, LLM's knowledge may not reflect recent or domain-specific facts; they could also exhibit hallucinations, which is difficult to audit or verify, leading to unfaithful and unsupported response. 
To address these limitations, knowledge-graph retrieval-augmented generation (KG-RAG) methods further extend LLMs with retrieved structural knowledge. Various studies such as ToG \cite{sun2023think}, GoG \cite{xu2024generate}, and CoK \cite{li2023chain} leverage chain-based KG-RAG approaches that performs reasoning over retrieved chains of KG facts. Specifically, they infer intermediate “bridge” entities at each hop, which repeatedly fetches triplets linked to the current entities from the KG, and
uses the LLM to choose which relation and entity to retain for the next step. These chain-based KG-RAG methods ground multi-step reasoning in explicit KG paths, thereby mitigating outdated knowledge, reducing hallucinations through factual grounding and rendering each inference step to be transparent.

Despite their strengths, existing chain-based KG-RAG methods exhibit two principal limitations as shown in Figure~\ref{fig:drawbacks_TOG}. First, they reason over the question in a holistic manner and lack a mechanism to structure the inference into targeted sub-tasks. In the upper half ``Lack of Plan'' example, one should first identify the notable person and then determine the nation they governed. Instead, ToG treats the query as an indivisible whole, leading it to select the superficially matching \textit{location.containedby} relation rather than the correct \textit{people.place\_live.person}.

Second, chain-based approaches cannot handle richer logical structures beyond simple, linear hops, such as conjunction questions that require the intersection of multiple constraint sets. Logic structure refers to the structural form of reasoning required to answer a question. In the lower half ``Lack of Logical'' example, the answer hinges on intersecting (a) countries bordering France and (b) countries with airports serving Nijmegen. ToG’s linear, single-path search prunes away additional candidates after the first hop, making it incapable of resolving the conjunction and thus yielding an incorrect result.

Motivated by these challenges, we propose the \textbf{P}redict–\textbf{D}ecompose–\textbf{R}etrieve–\textbf{R}eason (PDRR) framework. Inspired by pre-LLM semantic-parsing (SP-based) methods \cite{lan2022complex} that translate questions into executable KB logic forms, PDRR introduces a planning module (Predict, Decompose) for structuring multi-step inference, followed by a retrieval and reasoning module (Retrieve, Reason) that grounds the plan in KB facts and guides the LLM to execute it step by step, effectively simulating logical form execution over KGs. Finally, the question answering module utilizes the reasoning triples provided by the previous step to generate an answer.

First, our Predict stage classifies each query by type (i.e., chain or parallel) to determine an appropriate reasoning strategy and plan structure before retrieval begins. Next, the Decompose stage converts the question into a set of partial KG triples that reflect the plan, which breaks the overall query into manageable inference units, ensuring each reasoning step is focused and auditable. The Retrieve stage issues targeted KB lookups to fill in missing triple elements, which grounds the planned inference steps in factual knowledge. Finally, the Reason stage leverages the LLM to verify and complete each triple in accordance with the plan. By unifying logical-form-style planning with LLM-driven execution, PDRR supports a variety of reasoning patterns while preserving transparency.

We evaluate PDRR on four standard KBQA benchmarks: CWQ \cite{talmor-berant-2018-web}, WebQSP \cite{berant-etal-2013-semantic}, Simple Question\cite{bordes2015large}, and GrailQA\cite{gu2021beyond}. Experimental results show that our proposed KBQA model, PDRR, consistently outperforms existing methods across different LLM backbones and achieves superior performance on various types of questions, demonstrating the effectiveness of PDRR's explicit planning module and the precise control over retrieval and reasoning modules.

\section{Related Work} \label{sec:related_work}

\paragraph{Semantic Parsing-Based Methods}

Before LLMs, KBQA was dominated by SP-based approaches, which generate logical forms to query structured KBs.
CBR-KBQA \cite{das2021case} combines a non-parametric memory of question–logical form pairs with a parametric retriever to guide logical form generation.
TemplateQA \cite{zheng2018question} bypasses complex parsing by matching questions to a large set of binary templates.
SPARQA \cite{sun2020sparqa} uses a skeleton grammar and a BERT-based coarse-to-fine parser to handle complex questions.
While interpretable, SP-based methods are limited by incomplete KBs and the need for additional model training.

\paragraph{LLM Retrieval Augmented Methods}

The emergence of LLMs has led to LLM-only approaches that require no additional training and leverage internalized general knowledge to mitigate KB incompleteness, such as IO \cite{brown2020language}, CoT \cite{wei2022chain}, and SC \cite{wang2022self}.
Despite their simplicity, these methods struggle with outdated knowledge, limited reasoning transparency, and hallucinations.
To mitigate these issues, LLM+KG approaches have been developed.
ToG \cite{sun2023think} uses an LLM to guide reasoning over KGs by dynamically selecting relations and entities.
CoK \cite{li2023chain} adaptively selects KGs and performs stepwise retrieval.
GoG \cite{xu2024generate} augments KG coverage by letting LLMs infer missing links.

However, most chain-based KG-RAG methods lack explicit planning and are confined to chain reasoning, limiting reasoning control.
Recent methods like chatKBQA \cite{luo2023chatkbqa} and RoG \cite{luo2023reasoning} introduce planning by fine-tuning LLMs to generate logic forms or reasoning paths, but they remain training-intensive. PoG \cite{chen2024plan} and DoG \cite{li2025decoding} plan well but remain limited on non-chain complex questions.
To address this, we propose PDRR: a training-free framework with explicit planning, enabling precise reasoning control and handling of complex question structures.

\section{Preliminary} \label{preliminary}
In this section, we first introduce Knowledge Graphs (KGs). Then, we use notations of KGs to describe reasoning triples and Knowledge Base Question Answering (KBQA).

\paragraph{Knowledge Graphs}

A KG $\mathcal{G}$ consists of factual triples $(e^{h}, r, e^{t}) \in \mathcal{G} \subseteq \mathcal{E} \times \mathcal{R} \times \mathcal{E}$, where $e^{h}$, $e^{t}$ represent head and tail entities, and $\mathcal{E}$ and $\mathcal{R}$ denote the sets of entities and relations, respectively.

\paragraph{Reasoning Triples}

The Reasoning Triples $\mathcal{T}^q$ denote the set of triples retrieved from the KG $\mathcal{G}$ to answer the question $q$, formally defined as:
\begin{equation*}
\textstyle \mathcal{T}^q=\left\{\mathcal{T}_n^q=\left(e_n^h, r_n, e_n^t\right) \mid n=1,2,\ldots,m\right\} \subseteq \mathcal{G}
\end{equation*}
Here, $\mathcal{T}_n^q$ denotes the $n$-th reasoning triple.

%.....figure in the Approach
\begin{figure*}[t]
\centering
\includegraphics[width=2.12\columnwidth]{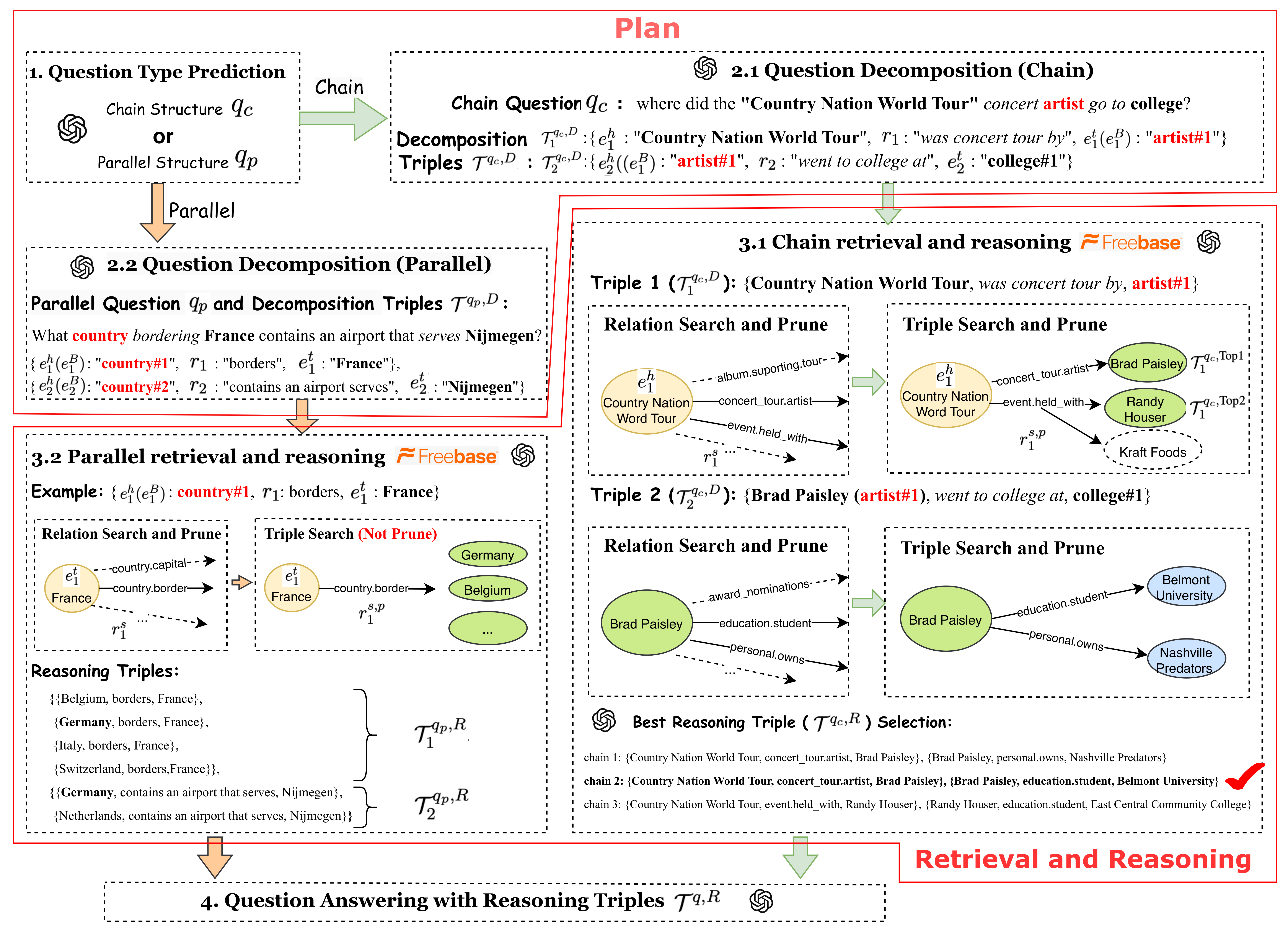}
\caption{The framework of the PDRR method. 
The process follows the Predict-Decompose-Retrieve-Reason pipeline. 
Dashed lines and circles indicate pruned components with low relevance to the specific decomposed triple $\mathcal{T}_{n}^{q,D}$. Labeled \textit{entity\#index} (e.g., \textit{artist\#1}) elements denote key bridge entities $e_{n}^{B}$ that are essential for reasoning.}
\label{fig:Overall_framework}
\end{figure*}

\paragraph{Knowledge Base Question Answering}

KBQA involves reasoning over a KG to answer natural language questions. Formally, given a question $q$ and a knowledge graph $\mathcal{G}$, the objective is to learn a function $f$ that returns answers $a \in \mathcal{A}_q$ based on the information in $\mathcal{G}$, i.e., $a = f(q, \mathcal{G})$.
In line with previous studies \cite{sun2019pullnet,jiang2022unikgqa}, we assume that the entities $e_q \in \mathcal{Q}_q$ mentioned in the question and the gold answers $a \in \mathcal{A}_q$ are annotated and linked to entities in the KG, such that $\mathcal{Q}_q, \mathcal{A}_q \subseteq \mathcal{E}$, where $\mathcal{E}$ denotes the entity set of $\mathcal{G}$.

\section{Approach}

We propose a training-free method to address the lack of planning and logical structure handling in existing chain-based KG-RAG approaches.
Our approach consists of three main modules: Plan, Retrieval and Reasoning, and Question Answering.
The Plan module predicts the question type (i.e., chain or parallel) and decomposes the question accordingly.
The Retrieval and Reasoning module retrieves the KB for relevant facts and employs the LLM to reason over and complete the decomposed triples.
Finally, the Question Answering Module generates the final answer.
Figure~\ref{fig:Overall_framework} illustrates the overall framework of our approach.

\subsection{Plan Module}
Our Plan module resembles logical form generation in SP-based methods. It first predicts the question type (i.e., chain or parallel) and then decomposes the question into decomposition triples, which serve a similar role to logical forms by guiding subsequent retrieval and reasoning.
The motivation is to support not only chain-structured questions but also more complex reasoning patterns, while enabling precise downstream control through the generated decomposition triples.
\subsubsection{Question Type Prediction}
Given a question $q$, we first use few-shot learning with LLMs predict its structural type, which we classify into two main categories: chain structure $q_c$ and parallel structure $q_p$.
The chain structure is the most common type in chain-based KG-RAG methods. Its reasoning process is sequential, where each step depends on the bridge entity $e_{n}^{B}$ identified in each step of reasoning triple $\mathcal{T}_n^q$. Therefore, the reasoning steps are interdependent and must be performed in order.
In contrast, the parallel structure comprises multiple logically independent sub-steps that can be executed concurrently. More detail refers to section: Discussion of Chain and Parallel Reasoning.

\subsubsection{Question Decomposition}

Based on the predicted question type, question $q$ is decomposed into KG-style triples $\mathcal{T}^{q,D}$,  providing precise control for downstream processing.

For chain-structured questions $q_c$, we identify all bridge entities $e_n^B$ (denoted as \textit{entity\#index}) to distinguish them in multi-hop reasoning. Each decomposition triple $\mathcal{T}{n}^{q_{c},D} = (e_n^h, r_n, e_n^t)$ connects entities via relation $r_n$, with adjacent triples linked through $e_n^B$ such that $e_{n+1}^h = e_n^t = e_n^B$, forming a reasoning chain.
For example, in Figure~\ref{fig:Overall_framework}, the bridge entity \textit{artist\#1} connects two steps: identifying the \textit{artist\#1} of Country Nation World Tour, and determining the college that the \textit{artist\#1} attended. 
%The corresponding decomposition triples are illustrated in the figure~\ref{fig:Overall_framework}.

For parallel-structured questions, we similarly identify bridge entities and construct KG-style triples along independent reasoning paths. Unlike chain structures, these steps are mutually independent and can be executed in parallel.
For example, in Figure~\ref{fig:Overall_framework}, the reasoning involves two conditions: identifying \textit{country\#1} bordering France and \textit{country\#2} with an airport serving Nijmegen, as reflected in the decomposition triples.

\subsection{Chain Retrieval and Reasoning Module}
In Retrieval and Reasoning module, we adopt different reasoning strategies based on the identified question type. Similar to how SP-based methods use logic forms to retrieve information from KBs, we leverages KB knowledge and employs the LLM as an agent to complete missing information in the decomposition triples $\mathcal{T}^{q,D}$ generated by the plan module. In this module, if no relevant information can be retrieved from the KB, the system directly generates the answer using the CoT approach.

% \subsubsection{Chain Retrieval and Reasoning}

Given a chain-type question $q_c$ and its decomposition triples $\mathcal{T}^{q_{c},D}$, we sequentially complete them by identifying bridge entities step by step, enabling accurate construction of reasoning triples and final answer derivation.

\subsubsection{Complete First Decomposition Triples}

We first extract $\mathcal{T}_{1}^{q_{c},D} = (e_1^h, r_1, e_1^t (e_1^B))$ and retrieve the bridge entity $e_1^B$ using the method illustrated in Figure~\ref{fig:Overall_framework}.

\begin{table*}[t]
\setlength{\tabcolsep}{1mm} 
\small 
\centering
\begin{tabular}{m{1.9cm}|p{5cm}|p{7.9cm}|p{2cm}}
\clineB{1-4}{2}
\textbf{Question type} & \textbf{Question Example} & \textbf{Canonical Reasoning Logic} & \textbf{Expected Reasoning Type} \\
\clineB{1-4}{1}
Composition & Where did the "Country Nation World Tour" concert artist go to college? 
& First, identify the artist of the concert tour "Country Nation World Tour," and then determine which college this artist attended.
& Chain \\
\clineB{1-4}{1} 
Conjunction & What country bordering France contains an airport that serves Nijmegen? 
& Find the intersection of two sets: (1) all countries that border France, and (2) all countries that contain an airport that serves Nijmegen.
& Parallel \\
\clineB{1-4}{2}
Comparative & Which of the countries bordering Mexico have an army size of less than 1050? 
& First, identify all countries bordering Mexico, and then select those with an army size of less than 1050.
& Parallel \\
\clineB{1-4}{2}
Superlative & What movies does taylor lautner play in and is the film was released earliest?
& First, identify all films in which Taylor Lautner appeared, and then find the one that was released the earliest.
& Parallel \\
\clineB{1-4}{2}
\end{tabular}
\caption{Examples of the four question types in the CWQ dataset, their corresponding canonical reasoning logic, and the expected reasoning type for each.}
\label{tab:cwq_question_type}
\end{table*}

%...............................Main results
\begin{table*}[t]
%\resizebox{1\textwidth}{!}{\renewcommand{\arraystretch}{1}
\setlength{\tabcolsep}{1.6mm} 
%\renewcommand{\arraystretch}{1.1}
%\small
    \centering
    \begin{tabular}{lccccccccc}
    \clineB{1-7}{2}
    \toprule
    \multirow{2}{*}{Method} & \multicolumn{5}{c}{CWQ} & \multicolumn{1}{c}{WebQSP} & \multicolumn{1}{c}{SimpleQA} & \multicolumn{1}{c}{GrailQA} \\
    \cmidrule(lr){2-6}  
    %\cmidrule(lr){7}
    %\cmidrule(lr){8}
                        & All & Composition & Conjunction & Comparative & Superlative & All & All & All \\
    \midrule
    \multicolumn{9}{c}{\textit{Without KB Knowledge, Without Training}} \\
     \clineB{1-10}{1}
        IO                    & 44.3  & 41.3 & 50.6 & 33.8 & 27.9 & 70.8 & 27.1 & 36.8   \\ 
        CoT                  & 44.8 & 44.2 & 49.5 & 30.5 & 27.4 & 72.1 & 27.3 & 37.2   \\
        %SC                   &  &  &  &  &  &    \\
        \textbf{PDR} (ours)     & 45.7 & 49.3 & 46.4 & 31.5 & 26.4 &  68.9 & 27.5 & 37.5 \\
    \midrule
    \multicolumn{9}{c}{\textit{With KB Knowledge, With Training}} \\
    \clineB{1-9}{1}
    SPARQA $^{\alpha}$  & 31.6 & - & - & - & - & - & - & -     \\
    UniKGQA  $^{\alpha}$  & 51.2 & - & - & - & - & 77.2 & - & -  \\
    RoG    $^{\alpha}$  & 62.6 & - & - & - & - & 85.7 & - & -  \\
    CBR-KBQA  $^{\alpha}$   & 67.1 & - & - & - & - & 69.9 & - & -  \\
    ChatKBQA  $^{\alpha}$  & 82.7 & - & - & - & - & 83.2 & - & - \\
    \midrule
    \multicolumn{9}{c}{\textit{With KB Knowledge, Without Training}} \\
    \clineB{1-9}{1}
        StructGPT $^{\alpha, \beta}$     & - & - & - & - & - & 72.6 & - & - \\
         \textbf{PDRR}(ours) $^{\beta}$ & 37.7 & 34.5 & 43.9 & 26.3 & 25.9 & 75.4 & - & - \\
        ToG                  & 48.9 & 49.9 & 50.1 & \underline{42.7} & \textbf{37.1} & \textbf{80.2} & \underline{57.2} & \underline{65.5}  \\
        \textbf{PDRR}(ours)        & \underline{59.6} & \textbf{66.2}  &\underline{59.1} & 38.5 & 34.5 & \underline{79.2} & \textbf{59.3} & \textbf{71.7}   \\ 
    \hdashline
        PDRR(gold type)(ours)    & \textbf{62.2} & \underline{65.5} &\textbf{64.6} & \textbf{43.7} & \underline{36.6} & - & - & -    \\ 
    \clineB{1-9}{2}
    \end{tabular}%}
\caption{
Hit@1 accuracy results with different baselines on KGQA datasets.
\textbf{Bold} denotes the best performance among the training-free methods, and \underline{underline} indicates the second-best.
Results marked with superscripts $\alpha$ are taken directly from the original papers.
Except for models marked with $\beta$, which use GPT-3.5-turbo, all other results are reproduced using GPT-4o as the backbone. 
Our proposed methods include PDR, PDRR, and PDRR (gold type).
In the gold type setting (an ablation setup), ground-truth question types from CWQ are used instead of LLM predictions. Composition-type questions are handled with chain reasoning, while all others use parallel reasoning.
}
    \label{tab:main_accuracy_results}
\end{table*}
%.............................Main results

\begin{itemize}
  \item \textbf{Relation Search and Prune}
We first apply SPARQL fuzzy matching to obtain the Freebase entity ID of the non-bridge entity in the triple thought its string.
In the \textbf{Search} phase, we query the KB with this entity ID to obtain all connected relations $r_1^s$.
In the case of Figure~\ref{fig:Overall_framework}, we first identify the entity ID of $e_1^h$( \textit{Country Nation World Tour}), and then retrieve all connected relations $r_1^s$.

We then \textbf{prune} the retrieved relation set $r_1^s$. Unlike ToG \cite{sun2023think}, which ranks relations $r_{1,i}^s$ by their similarity $Sim(r_{1,i}^s, q)$ to the entire question $q$ and may overlook local information, we rank them by their similarity $Sim(r_{1,i}^s, \mathcal{T}_{1}^{q_c,D})$ to the specific decomposition triple $\mathcal{T}{1}^{q_c,D}$. This yields the pruned set $r_1^{s,p}$.
This helps avoid errors such as the one in Figure~\ref{fig:drawbacks_TOG}, where ToG selects \textit{contained by}, which aligns with the global semantics of “\texttt{What nation}” instead of the correct local relation \textit{people.place\_live.person}.  

By aligning pruning with the decomposition triple, our approach ensures precise control and preserves step-level semantics.  As shown in the Figure~\ref{fig:Overall_framework}, we obtain a pruned relation set $r_{1}^{s,p}=$ \{\textit{concert\_tour.artist}, \textit{event.held\_with}\}, which is most similar to the decomposition triple $\mathcal{T}_{1}^{q_c,D}$.

  \item \textbf{Triple Search and Prune}
Given ID of $e_1^h$ and the pruned relation set $r_1^{s,p}$, we use SPARQL to \textbf{search} all tail entities (bridge entities) $e_1^t(e_1^B)$ to construct the candidate triple set $\mathcal{T}_{1}^{q_{c},\text{Cand}}$. In the Triple \textbf{Prune} stage, instead of pruning only entities as in ToG, which inherits the same limitations as its relation pruning, we rank candidate triples $\mathcal{T}_{1}^{q_{c},\text{Cand}_i}$ based on their similarity $Sim(\mathcal{T}_{1}^{q_{c},\text{Cand}_i}, \mathcal{T}_{1}^{q_c,D})$ to the decomposition triple $\mathcal{T}_{1}^{q_c,D}$ and retain the top two:
$\mathcal{T}_{1}^{q_{c},\text{Top1}} =$ \{\textit{Country Nation World Tour}, \textit{concert\_tour.artist}, \textit{Brad Paisley}\}, and $\mathcal{T}_{1}^{q_{c},\text{Top2}}$.

\end{itemize}

\subsubsection{Complete Rest Decomposition Triples}
Through Relation Search and Prune and Triple Search and Prune, we identify the top two bridge entities in first decompostion triple: $e_{1}^{B,\text{top1}}$ and $e_{1}^{B,\text{top2}}$ .
We then replace the bridge entity in next decompostion triple $\mathcal{T}_{2}^{q_{c},D}$ with the bridge entities from the previous step and repeat the same procedure until all decomposition triples in $\mathcal{T}^{q_{c},D}$ are completed.

\subsubsection{Choose Best Reasoning Triples}

We apply beam search to retain the top-2 triples at each hop, preserving sufficient information. For example, for a 2-hop question, this yields 4 reasoning triples. The LLM then selects the most relevant triple:$\mathcal{T}^{q_{c},R}$ for answering the question.

\subsection{Parallel Retrieval and Reasoning Module}

Parallel-structured questions $q_p$ follow a non-sequential reasoning logic, where the decomposition triples $\mathcal{T}^{q_{p},D}$ are mutually independent, allowing all retrieval and reasoning steps to be executed concurrently.

For example, given the first decomposition triple $\mathcal{T}_{1}^{q_{p},D}$ in Figure~\ref{fig:Overall_framework}, we perform relation search and pruning, followed by triple search, using the same procedure as in the chain setting.
Unlike the chain case, we skip triple pruning, as all reasoning triples are needed for the final answer. This process results in a set of reasoning triples $\mathcal{T}_1^{q_{p},R}$. 

We repeat this process for all decomposition triples $\mathcal{T}_k^{q_{p},D}$ to obtain the corresponding reasoning triples $\mathcal{T}_k^{q_{p},R}$, where $k$ indexes each decomposition triple.

\subsection{Discussion of Chain and Parallel Reasoning} \label{section:comparison_chain_reasoning}

While our question type classification is based on reasoning strategy, another key criterion is the number of bridge entities that the model should memorize during the reasoning process. Chain-structured questions require exactly one bridge entity between each pair of triples to support step-by-step reasoning, whereas parallel-structured questions allow multiple bridge entities without such constraints.

When applying chain reasoning to parallel-structured questions, retaining all retrieved triples without pruning can still lead to correct answers. For example, in Figure~\ref{fig:Overall_framework}, all \textit{countries} (bridge entities) bordering France are considered and individually checked for containing an airport that serves Nijmegen, which can lead to the right answer. However, this significantly increases computational cost. Introducing pruning in chain reasoning mitigates this cost but risks discarding correct answers. 

Therefore, parallel reasoning can be regarded as a complementary strategy to chain reasoning, particularly in cases with numerous bridge entities where computational efficiency becomes critical.

In practice, the logical structures or types of questions are not limited to just chain and parallel. 
%For example, SP-based methods often train models using the training set to generate logic forms that can handle more complex logical structures. 
In our work, we focus only on chain and parallel logic structures because a well-designed parallel structure that complements the chain structure, combined with the general knowledge and reasoning capabilities of LLMs, is sufficient to address the majority of KBQA questions. This also allows our method to remain training-free, avoiding additional computational costs.

\subsection{Question Answering Module}

Given the original question $q$, the decomposition triples $\mathcal{T}^{q,D}$, and the reasoning triples $\mathcal{T}^{q,R}$. The LLM is guided to answer the question $q$ by leveraging the information in retrieved reasoning triples $\mathcal{T}^{q,R}$ in accordance with the reasoning logic encoded in $\mathcal{T}^{q,D}$.

\section{Experiments}

\subsection{Experiment Setup}

\paragraph{Dataset}

To evaluate performance beyond simple chain-structured QA, we evaluate the model on the CWQ \cite{talmor-berant-2018-web} test set (3,531 questions), which includes four question types: composition (45\%), conjunction (45\%), comparative (5\%), and superlative (5\%).
%Composition questions are expected to follow chain reasoning, while the other three suit parallel reasoning. 
Table~\ref{tab:cwq_question_type} shows examples, reasoning logic, and expected types.
For additional validation, we evaluate the model on the test sets of WebQSP \cite{berant-etal-2013-semantic} (1,639 questions, CC License), SimpleQuestions \cite{bordes2015large} (CC License), and GrailQA \cite{gu2021beyond}. 
%These datasets primarily consist of chain-structured questions involving one- or two-hop reasoning.
%Notably, for fair comparison, we directly use the 1,000 test questions randomly selected in the ToG \cite{sun2023think} from the full SimpleQuestions and GrailQA test sets.

%.....................................Different Backbone
\begin{table}[t]
\setlength{\tabcolsep}{1.6mm} 
%\renewcommand{\arraystretch}{1}
%\small
%\resizebox{1\textwidth}{!}{\renewcommand{\arraystretch}{1}
   \centering
   \begin{tabular}{lccccc}
   \clineB{1-6}{2}
   \toprule
    \multirow{2}{*}{Method} & \multicolumn{5}{c}{CWQ} \\
    \cmidrule(lr){2-6}
     & All & Compo & Conju & Compa & Super  \\
   \midrule
   \multicolumn{6}{c}{\textit{GPT-3.5-turbo-0125}} \\
   \clineB{1-6}{1}
   CoT      & 36.8 & 33.8 & 43.0 & \underline{26.8} & 20.8 \\
   ToG      & 30.9 & 30.1 & 35.6 & 16.4 & 15.8 \\
   PDRR    & \underline{37.7} & \underline{34.5} & \underline{43.9} & 26.3 & \underline{25.9}\\
   PDRR(gt)     & \textbf{40.7} & \textbf{36.1} & \textbf{48.6} & \textbf{29.6} & \textbf{26.4} \\
   \midrule
   \multicolumn{6}{c}{\textit{Llama3.3-Instruct}} \\
   \clineB{1-6}{1}
   CoT      & 46.0 & 41.5 & 52.4 & \underline{42.7} & 33.5   \\
   %ToG      &  &   \\
   PDRR    & \underline{54.1} & \underline{54.9} & \underline{57.7} & 36.2 & \underline{38.1}  \\
   PDRR(gt)    & \textbf{58.7} & \textbf{56.4} & \textbf{65.0} & \textbf{46.5} & \textbf{39.1}  \\
   \midrule
   \multicolumn{6}{c}{\textit{Deepseek-V3}} \\
   \clineB{1-6}{1}
   CoT      & 44.3 & 43.0 & 48.5 & 35.7 & 29.9  \\
   ToG      & 48.0 & \underline{48.5} & 51.7 & 35.7 & 27.4  \\
   PDRR          & \underline{56.0} & \textbf{60.3} & \underline{57.0} & \underline{41.3} & \underline{30.5}   \\
   PDRR(gt)      & \textbf{57.1} & \textbf{60.3} & \textbf{57.5} & \textbf{43.7} & \textbf{42.6}   \\
   \midrule
   \multicolumn{6}{c}{\textit{GPT-4o-2024-11-20}} \\
   \clineB{1-6}{1}
   CoT      & 44.8 & 44.2 & 49.5 & 30.5 & 27.4   \\
   ToG      & 48.9 & 49.9 & 50.1 & \underline{42.7} & \textbf{37.1}   \\
   PDRR       & \underline{59.6} & \textbf{66.2} & \underline{59.1} & 38.5 & 34.5 \\
   PDRR(gt)      & \textbf{62.2} & \underline{65.5} & \textbf{64.6} & \textbf{43.7} & \underline{36.6}   \\
   \clineB{1-6}{2}
   \end{tabular}
   %}
\caption{Performance of PDRR with different backbone models on overall accuracy and by question type in the CWQ dataset. ‘gt’ denotes the gold type setting.}
\label{tab:different_Backbone}
\end{table}
%....................................Different Backbone

\paragraph{Evaluation Metrics}
Consistent with prior studies \cite{luo2023reasoning,sun2023think}, we adopt Hits@1 as the evaluation metric, which reflects the percentage of questions for which the top-1 rank predicted answer is correct.

\paragraph{Baseline}
We divide the baselines into three groups. 
The first group includes LLM-only methods including IO \cite{brown2020language} and CoT \cite{wei2022chain}, and our ablated variant PDR (Predict-Decompose-Reason), which removes the retrieve stage from PDRR.
The second group incorporates external KBs and require additional training, including UniKBQA \cite{jiang2022unikgqa}, ROG \cite{luo2023reasoning}, CBR-KBQA \cite{das2021case}, SPARQA \cite{sun2020sparqa}, and ChatKBQA \cite{luo2023chatkbqa}.
The third group leverages external KBs without additional training, such as ToG \cite{sun2023think}, StructGPT \cite{jiang2023structgpt}, and our model, PDRR.

\paragraph{Implementation}

For all intermediate steps, the max token length is 256, and 1024 for final answer generation. A temperature of 0.1 is used to reduce hallucinations and improve control.
We adopt 5-shot prompting for answer generation and question type classification, and 3-shot for all other components.

\subsection{Accuracy Result}

\subsubsection{Main Results}

In this section, we compare PDRR with various baselines across KGQA datasets as shown in Table~\ref{tab:main_accuracy_results}.
On the more challenging CWQ dataset, which features diverse question structures beyond simple chains, PDRR performs competitively with training-based methods and outperforms training-free ones like ToG by nearly 10\% in accuracy.

By question type, PDRR significantly surpasses ToG on composition questions (66.2 vs. 49.9), highlighting the effectiveness of the Plan Module in guiding step-by-step reasoning. It also achieves higher accuracy on conjunction questions and remains competitive on comparative and superlative types, showing that augmenting the Retrieval and Reasoning Module with complementary parallel reasoning enhances performance on complex, diverse questions. 

On the simpler WebQSP, SimpleQuestions, and GrailQA datasets, which contain only composition-type questions, ToG and PDRR achieve comparable performance on WebQSP, while PDRR consistently outperforms ToG on both GrailQA and SimpleQuestions, indicating stronger generalizability across both simple and complex cases.

\subsubsection{Different Backbone Models}

To evaluate the robustness of PDRR, we test its performance on the CWQ dataset using different LLM backbones to assess its effectiveness beyond GPT-4o.
Specifically, we compare CoT, ToG, PDRR, and PDRR (gt) across four models: GPT-3.5-turbo, DeepSeek-V3, and GPT-4o. Additionally, we evaluate CoT, PDRR, and PDRR (gt) on LLaMA-3.3B-Instruct.
As shown in Table~\ref{tab:different_Backbone}, PDRR consistently outperforms across all models. The performance gap is modest on GPT-3.5-turbo but becomes more significant on the stronger LLMs.
These results confirm that PDRR is robust and not dependent on any specific language model.

\subsection{Discussion of Question Structure}
A core component of PDRR is the use of the Plan Module to predict the question structure type and apply corresponding decomposition and reasoning strategies. Thus, investigating this process is essential.

%.....figure in the question type prediction
\begin{figure}[t]
\centering
\includegraphics[width=1\columnwidth]{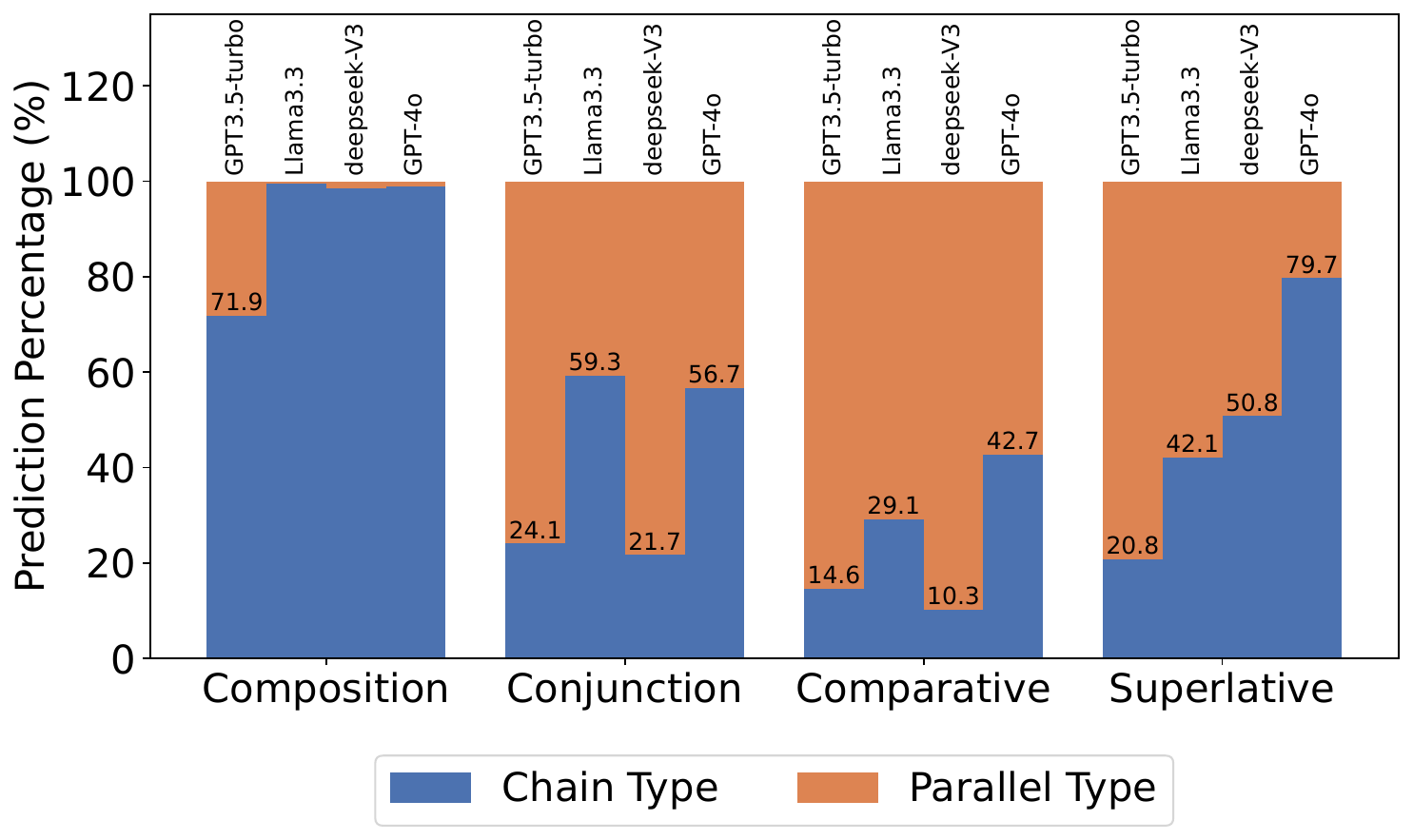}
\caption{Predicted question structure types (chain or parallel) by different LLMs on various question types in the CWQ dataset. Blue indicates that the LLM predicts the question as a chain structure, while orange indicates a parallel structure.}
\label{fig:question_type_prediction}
\end{figure}

%......figure in the question type prediction

%.....figure in the question type prediction
\begin{figure}[t]
\centering
\includegraphics[width=1\columnwidth]{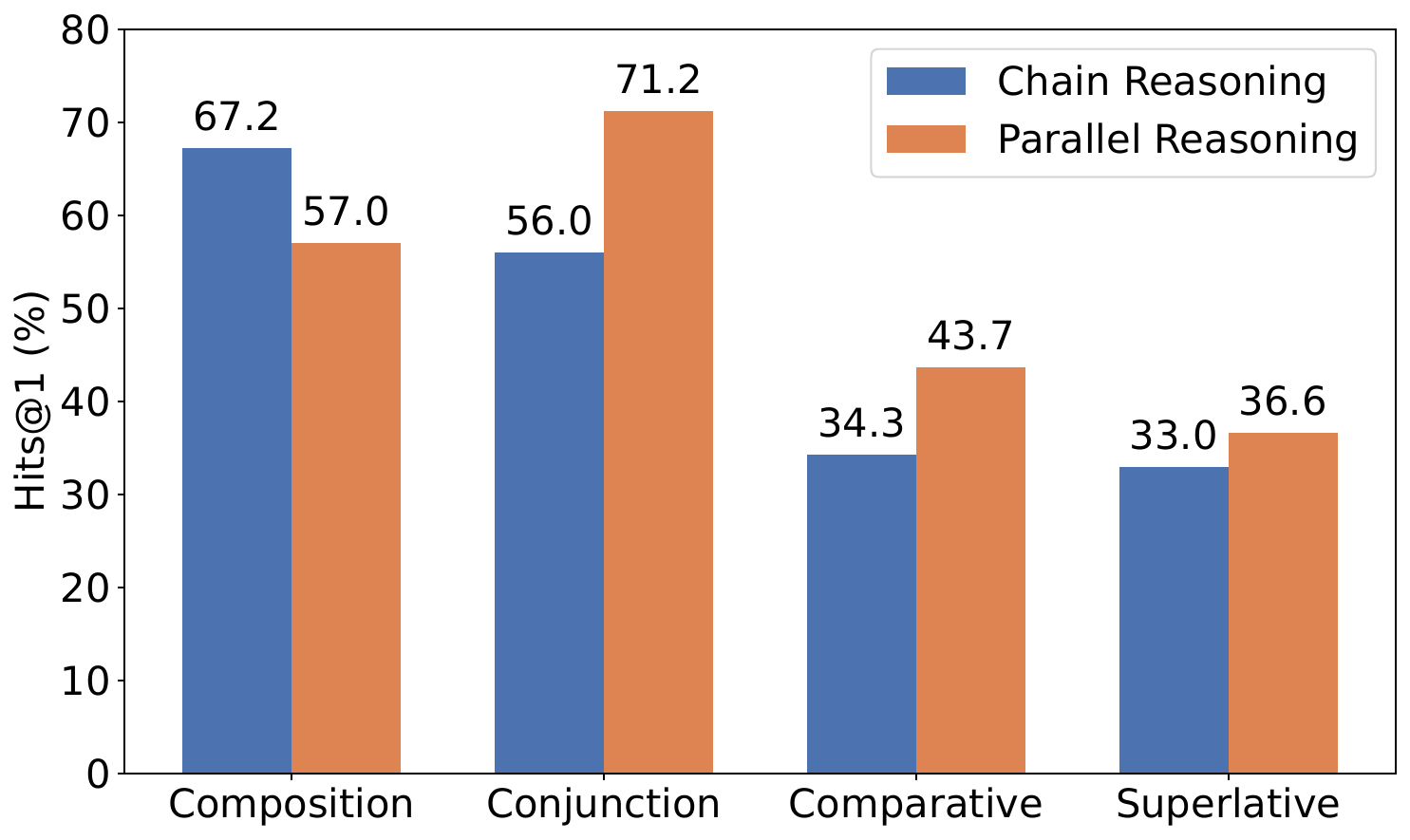}
\caption{Hits@1 accuracy of different question types in the CWQ dataset under chain and parallel reasoning strategies. The evaluation includes 500 chain, 500 intersection, 213 comparative, and 197 superlative questions. GPT-4o is used as the LLM backbone.}
\label{fig:reasoning_accuracy}
\end{figure}

%......figure in the question type prediction

%.....figure in the question type prediction
\begin{figure}[t]
\centering
\includegraphics[width=1\columnwidth]{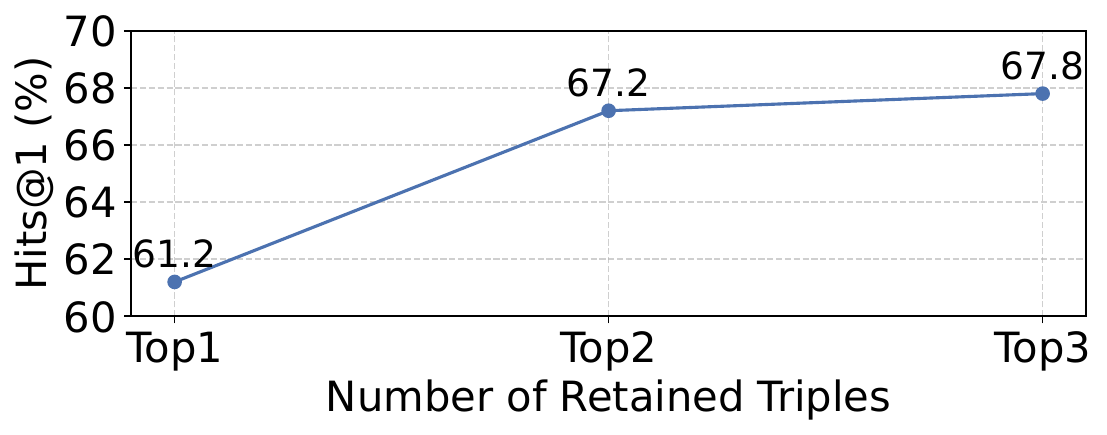}
\caption{Hits@1 accuracy on the first 500 composition-type questions in CWQ using chain reasoning with different number of retained triples.}
\label{fig:ablation_search_width}
\end{figure}

\paragraph{Question Type Prediction} \label{subsec:question_structure_prediction}

We analyze how different LLMs predict the structure type (chain or parallel) for various CWQ question types, as shown in Figure~\ref{fig:question_type_prediction}. 

For composition-type questions, most LLMs correctly predict a chain structure, consistent with expectations. For conjunction-type questions, which are best handled by parallel reasoning, GPT-3.5-turbo and DeepSeek-V3 predict the correct structure in most cases, while Llama3.3-Instruct and GPT-4o achieve a much lower rate. Despite this, the results in Table~\ref{tab:main_accuracy_results} reveal that many misclassified conjunction questions still lead to correct answers, as they can be effectively addressed by either chain or parallel reasoning. 

This phenomenon can be explained by the number of bridge entities: when only one is involved, both chain and parallel reasoning are generally effective; with more, parallel reasoning becomes notably more robust. This supports the discussion in Section: Discussion of Chain and Parallel Reasoning. 

%Appendix includes two detailed cases illustrating both strategies in single bridge-entity scenarios, along with additional analyses of comparative and superlative questions. 

%......figure in the question type prediction

\paragraph{Different Reasoning Strategies}

We evaluate the effectiveness of chain and parallel reasoning across four CWQ question types using Hits@1 accuracy, as shown in Figure~\ref{fig:reasoning_accuracy}. Chain reasoning performs significantly better than parallel on composition-type questions (67.2\% vs. 57.0\%), while parallel reasoning clearly outperforms chain on conjunction-type questions (71.2\% vs. 56.0\%) and slightly outperforms it on comparative and superlative questions.

These results align with our hypothesis: composition-type questions favor chain reasoning, whereas the others benefit more from parallel reasoning. In contrast, methods like ToG, which rely solely on chain reasoning, underperform on non-chain questions. This highlights the need to adapt reasoning strategies to question type.

\subsection{Ablation Study}

\paragraph{Individual effects of the Retrieval and Reasoning Module}
We conducted experiments where this module was removed, relying only on the Plan Module and the final QA step to generate answers. The results are shown in Table~\ref{tab:main_accuracy_results}  under “PDR (ours)”, where the performance on CWQ is 45.7\%. This already outperforms other prompting-based baselines such as IO (44.8\%) and CoT (44.3\%), demonstrating the effectiveness of the Plan Module. At the same time, the significant performance drop from the full PDRR (59.6\%) to PDR (45.7\%) confirms that the Retrieval and Reasoning Module plays a crucial role in the overall framework.

\paragraph{Number of Retained Triples Within Chain Reasoning}

%......figure in the question type prediction
We aim to explore the impact of the Number of Retained Triples within Chain Reasoning on the experimental results.
As shown in Figure~\ref{fig:ablation_search_width}, increasing the number of retained triples from top-1 to top-2 significantly improves performance (61.2\% to 67.2\%). Although top-3 offers a minor additional gain (+0.6\%), we adopt top-2 to balance accuracy and computational cost.
%We add more ablation study and case study in Appendix.

\section{Conclusion}

To address the limitation of chain-based KG-RAG methods, which are restricted to simple chain-structured questions due to the lack of planning and logical structuring.
We propose PDRR, which first \textbf{predicts} the question type and \textbf{decomposes} the question into structured triples. Then \textbf{retrieves} relevant information from KBs and guides the LLM as an agent to \textbf{reason} over and complete the decomposed triples.
Experimental results show that PDRR consistently outperforms existing methods across various LLM backbones, with up to a 10\% gain on CWQ using GPT-4o. It also performs robustly across diverse question types. 
%Additionally, our in-depth analysis of question structures reveals that LLMs rely not only on the structural form of the question but also on the number of bridging entities they must retain during reasoning.

\section{Acknowledgment}

This work was partially supported by JST SPRING JPMJSP2110 (YZ), JSPS KAKENHI 22H05106, 23H03355, and JST CREST JPMJCR21N3 (HS).

\bibliography{aaai2026}

% Check whether the conference requires a reproducibility checklist to be included in the paper.
% If so, you can uncomment the following line and ajust the path to include it.
% \input{../../ReproducibilityChecklist/LaTeX/ReproducibilityChecklist.tex}

\clearpage

%.............................Appendix
\appendix
\section{Appendix}

\subsection{Experiment Detail}

\subsubsection{Experiment Dataset Detail}

The CWQ dataset mainly consists of complex-structured questions, whereas the other three datasets—WebQSP, SimpleQuestions, and GrailQA—primarily contain chain-structured questions involving one- or two-hop reasoning.
For fair comparison, we use the same 1,000 test questions randomly selected in ToG \cite{sun2023think} from the full test sets of SimpleQuestions and GrailQA.

Meanwhile, all four datasets were designed for KBQA task, and we employ them for KBQA tasks, and all datasets have no individual people or offensive content.

\subsubsection{Experiment Implementation Deatil} \label{appendix:implementation}
We use four LLMs as backbones in our experiments: GPT-3.5-turbo (gpt-3.5-turbo-0125), GPT-4o (gpt-4o-2024-11-20), LLaMA3.3-70B-Instruct\footnote{https://www.llama.com.}, and DeepSeek-V3. GPT-3.5-turbo and GPT-4o are accessed via the OpenAI API\footnote{https://openai.com.}, while DeepSeek-V3 is accessed through the DeepSeek API\footnote{https://www.deepseek.com}.
 All experiments and datasets are conducted using Freebase \cite{bollacker2008freebase} as the underlying knowledge base.

Furthermore, the training of models was carried out on four A100 GPUs for the Llama3.3-Instruct inference task. Specifically, for the PDRR model, the running durations were roughly 120 hours for the CWQ dataset, 30 hours for WebQSP. For a fair comparison with ToG, we evaluate both models using the OpenAI API on the CWQ dataset. PDRR requires approximately 17 hours and costs around \$80, while ToG takes about 15 hours and costs \$75. On simpler datasets such as SimpleQuestions, PDRR completes in roughly 3.5 hours, compared to 4.5 hours for ToG. These results indicate that PDRR is not more resource-intensive than ToG in practice, with overall time and cost remaining reasonable and dependent on task complexity.

For the pruning stage in the two reasoning strategies, we adopt different approaches.
In chain reasoning, we retain only the top two most similar relations and two triples in each of the two pruning phases.
In parallel reasoning, after relation pruning, we do not prune the retrieved entities, which can result in hundreds of triples. From these, we randomly select 20 triples for the next reasoning stage.
Based on empirical statistics over 100 chain and 100 parallel reasoning samples, the token consumption of parallel reasoning is approximately 1.34 times higher than that observed in chain reasoning.

Our experiments were facilitated by leveraging PyTorch\footnote{https://pytorch.org},  Huggingface\footnote{https://huggingface.co}, and Numpy\footnote{https://numpy.org} as essential tools. Furthermore, We use ChatGPT\footnote{https://chat.openai.com} in our paper writing and programming. Finally, we obtain results by using a single run for all results.

\subsubsection{Question type details of CWQ dataset} \label{appendix:cwq_question_type}

Table~\ref{tab:cwq_question_type} provides examples of the four question types in the CWQ dataset, along with their corresponding canonical reasoning logic and the expected reasoning type for each.

From the reasoning logic, we observe that Composition questions align with standard chain reasoning, which can also be effectively handled by chain-based methods. In contrast, Conjunction questions are expected to be addressed using parallel reasoning.

The main challenge lies in classifying Comparative and Superlative questions. Logically, they follow a chain reasoning pattern, as identifying bridge entity is a prerequisite for the next reasoning step, and chain reasoning typically retains only one such entity. However, given the often large number of bridge entities involved, parallel reasoning becomes necessary to reduce computational cost while preserving relevant information—an issue we previously discussed in the Comparison of Chain and Parallel Reasoning section (Discussion of Chain and Parallel Reasoning).
Considering that our approach leverages LLMs instead of relying solely on KBs as in traditional SP-based methods, we opt for parallel reasoning to retain as many bridge entities as possible. The subsequent reasoning is then delegated to the LLM, which can effectively handle this complexity.

\subsection{Result Detail}

\subsubsection{Detail case example of Chain and Parallel reasoning type for Conjunction question} \label{appendix:chain_parallel_for_conjunction}

Table~\ref{tab:case_chain_and_parallel_for_conjunction} presents two representative examples to illustrate under what conditions conjunction questions can be successfully answered using either chain or parallel reasoning.

In the first example, the correct reasoning logic is to find the intersection of two sets: (1) regions where \textit{William Morris} serves as the religious head, and (2) regions that are part of \textit{the United Kingdom}. This clearly aligns with a parallel reasoning structure. However, when using chain reasoning, even though the second set contains multiple bridge entities, the first set includes only a single bridge entity—\textit{Wales}. As shown in the reasoning triples in the table, chain reasoning can still correctly answer the question.

In the second example, although the second set of bridge entities—countries in \textit{Eastern Europe} — includes many candidates, the first set, referring to the countries that appointed \textit{Mikheil Saakashvili} to a governmental position, contains only one entity: \textit{Georgia}, which is also the correct answer. Thus, chain reasoning can be used to answer the question as well.

These two examples demonstrate that for conjunction-type questions, if one reasoning chain (independent of the others) contains only a single bridge entity, the question can be answered using either parallel or chain reasoning. This supports the explanation of experimental results discussed in Section~\ref{subsec:question_structure_prediction}.

%.....................................ablation sentence or triples
\begin{table}[t]
%\resizebox{1\textwidth}{!}{\renewcommand{\arraystretch}{1.1}
\renewcommand{\arraystretch}{1.2}
   \centering
   \begin{tabular}{lccccc}
   \clineB{1-6}{2}
   {Methods} & {All} & {Compo} & {Conju} & {Compa} & {Super} \\
   \clineB{1-6}{2}
   Sentence & 63.4 & 64.1 & 68.8 & 30.0 & 41.0\\
   Triple   & 66.8 & 67.3 & 73.2 & 34.0 & 35.9\\
   \clineB{1-6}{2}
   \end{tabular}
   %}
\caption{Hits@1 accuracy on the CWQ dataset when generating answers using the LLM by either directly inputting the reasoning triples or first converting them into natural language sentences. The evaluation is conducted on the first 1000 questions.}
\label{tab:ablation_sentence_triples}
\end{table}
%...................................ablation sentence or triples

\begin{table*}[t]
%\resizebox{1\textwidth}{!}{\renewcommand{\arraystretch}{1.2}
\small
\setlength{\tabcolsep}{1.6mm} 
\renewcommand{\arraystretch}{1.2}
\centering
\begin{tabular}{p{1.5cm}|p{7.5cm}|p{7.5cm}}
\clineB{1-3}{2}
\textbf{Question} & \multicolumn{2}{l}{William Morris is religions head in which region that is part of the United Kingdom? (question type: \textbf{Conjunction})} \\
\clineB{1-3}{1}
\textbf{Reasoning Logic} & \multicolumn{2}{p{15cm}}{Find the intersection of two sets:(1) regions where William Morris serves as the religious head, and (2) regions that are part of the United Kingdom.} \\
\clineB{1-3}{1}
\textbf{Answer} & \multicolumn{2}{l}{Wales} \\
\clineB{1-3}{2}
\textbf{Reasoning type} & \textbf{Reasoning Triples} & \textbf{Result} \\
\clineB{1-3}{1}
Chain & $\mathcal{T}_1^{q_{c},R}$: [UnName Entity, religion.religious-organization-leadership.leader, William Morris], [UnName Entity, religion.religious-organization-leadership.jurisdiction, \textbf{Wales}], 

$\mathcal{T}_2^{q_{c},R}$ : [\textbf{Wales}, location.location.containedby, United Kingdom] &
Step 1: Identify the region where William Morris is a religious head. According to the retrieved reasoning chains, William Morris is associated with an unnamed religious entity that has jurisdiction over \textbf{Wales}.

Step 2: Confirm that Wales is part of the United Kingdom. The retrieved reasoning chains confirm that \textbf{Wales} is contained within the United Kingdom.\\
\clineB{1-3}{1} 
Parallel & $\mathcal{T}_1^{q_{p},R}$: [William Morris, religion.religious-leader.religious-leadership, UnName Entity], [\textbf{Wales}, religion.religious-leadership-jurisdiction.leader, UnName Entity]

$\mathcal{T}_2^{q_{p},R}$: [\textbf{Wales}, location.location.containedby, United Kingdom], [England, location.location.containedby, United Kingdom], [Northern Ireland, location.location.containedby, United Kingdom], [Scotland, location.location.containedby, United Kingdom] & 
Step 1: Identify the region where William Morris is a religious head. based on the retrieved triples, \textbf{Wales} is the region linked to religious leadership.

Step 2: Identify regions that are part of the United Kingdom. Based on the retrieved triples, the United Kingdom consists of England, Scotland, \textbf{Wales}, and Northern Ireland.

Step 3: Combine the information. Since \textbf{Wales} is part of the United Kingdom and is associated with religious leadership, it is the region where William Morris is likely the religious head.
\\

\clineB{1-3}{2}
\textbf{Question} & \multicolumn{2}{l}{What location that appointed Mikheil Saakashvili to governmental position is a country in Eastern Europe? } \\
\clineB{1-3}{1}
\textbf{Reasoning Logic} & \multicolumn{2}{p{15cm}}{Find the intersection the following two sets: (1) the countries that appointed Mikheil Saakashvili to governmental position, (2) the countries in Eastern Europe} \\
\clineB{1-3}{1}
\textbf{Answer} & \multicolumn{2}{l}{Georgia} \\
\clineB{1-3}{2}
\textbf{Reasoning type} & \textbf{Reasoning Triples} & \textbf{Result} \\
\clineB{1-3}{1}
Chain & Reasoning step 1: 

[UnName Entity, government.government-position-held.appointed-by, Mikheil Saakashvili], [UnName Entity, government.government-position-held.jurisdiction-of-office, \textbf{Georgia}] 

Reasoning step 2: 

[\textbf{Georgia}, location.location.containedby, Eastern Europe] & 
Step 1: Identify the location that appointed Mikheil Saakashvili to a governmental position. According to the retrieved reasoning chain, the jurisdiction of the office held by Mikheil Saakashvili is \textbf{Georgia}.

Step 2: Determine if Georgia is a country in Eastern Europe. According to the retrieved reasoning chain, \textbf{Georgia} is contained within Eastern Europe.\\
\clineB{1-3}{1} 
Parallel & Triple group 1: 

[UnName Entity, government.government-position-held.appointed-by, Mikheil Saakashvili], [UnName Entity, government.government-position-held.jurisdiction-of-office, \textbf{Georgia}] 

Triple group 2: 

[Eastern Europe, location.location.partially-contains, Atyrau Region], [Eastern Europe, location.location.partially-contains, \textbf{Georgia}], ...\textit{(other triples)}
& 
Step 1: Identify the location that appointed Mikheil Saakashvili to a governmental position. According to the retrieved triples, Mikheil Saakashvili was appointed by an unnamed entity, and the jurisdiction of office is \textbf{Georgia}.

Step 2: Determine the region that is located in Eastern Europe. According to the retrieved triples, \textbf{Georgia}, the Atyrau Region in Kazakhstan, and many other regions are geographically located in Eastern Europe.

Step 3: Find the intersection of the two sets. The final answer is \textbf{Georgia}.
\\
\clineB{1-3}{2}
\end{tabular}
%}
\caption{Case study of Conjunction-type question that can be correctly answered using either chain or parallel reasoning. "Reasoning Logic" in the table denotes standard reasoning logic for the question.}
\label{tab:case_chain_and_parallel_for_conjunction}
\end{table*}

\subsubsection{Experiment explanation and case study on Comparative and Superlative type question} \label{appendix:comparative_superlative}

First, as shown by the main accuracy results for each question type in Table~\ref{tab:main_accuracy_results}, both baseline methods and our PDRR perform significantly worse on comparative and superlative questions compared to composition and conjunction questions. This is due to the higher structural complexity of the former two types. As indicated by the canonical reasoning logic in Table~\ref{tab:cwq_question_type}, comparative and superlative questions require not only the sequential reasoning process of chain reasoning, but also the ability to handle a larger number of bridge entities, as provided by parallel reasoning.

Nevertheless, we favor using parallel reasoning for these two question types, as it enables maximal retention of the retrieved bridge entities, which are essential for subsequent reasoning steps. In contrast, chain reasoning reduces computational cost by retaining only one bridge entity at each step, often at the expense of losing key information. This hypothesis is supported by the results in Figure~\ref{fig:reasoning_accuracy}, which show that parallel reasoning outperforms chain reasoning on both comparative and superlative questions.

To further validate this, we provide case studies in Table~\ref{tab:appendix_case_comparative_superlative_1} and ~\ref{tab:appendix_case_comparative_superlative_2}. For the comparative question, parallel reasoning successfully retrieves and retains all relevant bridge entities (\texttt{Italy}, \texttt{Lazio}, \texttt{Province of Roma}), allowing for accurate comparison and judgment in the next step. However, chain reasoning retains only one entity (\texttt{Italy}), losing the other two critical bridge entities. A similar issue arises in the superlative question, where chain reasoning discards key information, while parallel reasoning preserves the complete context.

In summary, for comparative and superlative question types, parallel reasoning is the preferred approach when choosing between chain and parallel reasoning.

\begin{table*}[t]
%\resizebox{1\textwidth}{!}{\renewcommand{\arraystretch}{1.2}
\small
\setlength{\tabcolsep}{1.6mm} 
\renewcommand{\arraystretch}{1.1}
\centering
\begin{tabular}{p{1.5cm}|p{3.5cm}|p{11.5cm}}

\clineB{1-3}{2}
\textbf{Question} & \multicolumn{2}{l}{Where with a population once of less than 5732212 is Rome, Italy located?} \\
\clineB{1-3}{1}
\textbf{Type} & \multicolumn{2}{l}{Comparative} \\
\clineB{1-3}{1}
\textbf{Reasoning Logic} & \multicolumn{2}{p{15cm}}{First, identify all regions that contains Roma, and then select those with a population once less than 5732212.} \\
\clineB{1-3}{1}
\textbf{Answer} & \multicolumn{2}{l}{Province of Rome} \\
\clineB{1-3}{2}
\textbf{Reasoning type} & \textbf{Reasoning Triples} & \textbf{Result} \\
\clineB{1-3}{1}
Chain & 
Reasoning step 1: 

[Rome, location.location.containedby, Italy]

Reasoning step 2: 

[Italy, location.location.containedby, Europe]
& 
Step 1: Identify where Rome, Italy, is located. According to the retrieved reasoning chain, Rome is located in Italy. Additionally, Rome is part of the Lazio region, which is located on the Italian Peninsula and in Southern Europe.

Step 2: Determine if the location had a population of less than 5,732,212. Based on the reasoning chain, Italy, Lazio, and the Italian Peninsula have populations that could meet this criterion at certain points in history. \underline{However, the question does not specify a time frame}, so the most relevant answer is the broader location of Italy.

\{\underline{Italy}\}
\\
\clineB{1-3}{1} 
Parallel & 
Triple group 1: 

[Rome, location.location.containedby, Italy], [Rome, location.location.containedby, Lazio], [Rome, location.location.containedby, Province of Rome]
& Following the question decomposition triples:

Step 1: Identify where Rome, Italy is located. According to the retrieved triples, Rome is located in Italy, Lazio, and the Province of Rome.

Step 2: Identify which of these locations had a population of less than 5,732,212. Since the retrieved triples do not provide population data, I will rely on my own knowledge. The population of Italy is significantly higher than 5,732,212, so it does not qualify. Lazio, the region containing Rome, has a population higher than 5,732,212 as well. However, the Province of Rome (now part of the Metropolitan City of Rome) had a population of less than 5,732,212 at one point.

Step 3: Based on this reasoning, the location with a population once less than 5,732,212 where Rome is located is the Province of Rome.

\{\textbf{Province of Rome}\}
\\

\clineB{1-3}{2}
\end{tabular}
%}
\caption{First case study of comparative and superlative type questions. \underline{Underline} text indicates incorrect parts.}
\label{tab:appendix_case_comparative_superlative_1}
\end{table*}

\begin{table*}[t]
%\resizebox{1\textwidth}{!}{\renewcommand{\arraystretch}{1.2}
\small
\setlength{\tabcolsep}{1.6mm} 
\renewcommand{\arraystretch}{1.3}
\centering
\begin{tabular}{p{1.5cm}|p{6cm}|p{9cm}}

\clineB{1-3}{2}
\textbf{Question} & \multicolumn{2}{l}{Which college attended by Tennessee Williams has the largest population of postgraduates?} \\
\clineB{1-3}{1}
\textbf{Type} & \multicolumn{2}{l}{Superlative} \\
\clineB{1-3}{1}
\textbf{Reasoning Logic} & \multicolumn{2}{p{15cm}}{First, identify all the colleges that Tennessee Williams attended, and then find the one that has the largest number of postgraduates.} \\
\clineB{1-3}{1}
\textbf{Answer} & \multicolumn{2}{l}{University of Iowa} \\
\clineB{1-3}{2}
\textbf{Reasoning type} & \textbf{Reasoning Triples} & \textbf{Result} \\
\clineB{1-3}{1}
Chain & 
Reasoning Chain 1:   

[UnName\_Entity, education.education.student, Tennessee Williams], [UnName\_Entity, education.education.institution, University of Missouri], [University of Missouri, education.educational\_institution.campuses, University of Missouri]

Reasoning Chain 2:  

[UnName\_Entity, education.education.student, Tennessee Williams], [UnName\_Entity, education.education.institution, Washington University in St. Louis], [Washington University in St. Louis, education.educational\_institution.campuses, Washington University in St. Louis]
& 
Step 1: Identify the colleges attended by Tennessee Williams. According to the retrieved reasoning chain, Tennessee Williams attended Washington University in St. Louis and the University of Missouri.

Step 2: Determine the postgraduate population of each college. The reasoning chain does not provide explicit information about the postgraduate populations of these institutions. Using my own knowledge:

- Washington University in St. Louis is known for having a significant postgraduate population due to its strong graduate programs.

- The University of Missouri also has a notable postgraduate population, but it is generally smaller compared to Washington University in St. Louis.

Step 3: Compare the postgraduate populations of the two colleges. Based on my knowledge, Washington University in St. Louis has a larger postgraduate population than the University of Missouri.

\{\underline{Washington University in St. Louis}\}
\\
\clineB{1-3}{1} 
Parallel & 
Triple Groups:

[UnName\_Entity, education.education.student, Tennessee Williams], [UnName\_Entity, education.education.institution, Washington University in St. Louis] 

[UnName\_Entity, education.education.student, Tennessee Williams], [UnName\_Entity, education.education.institution, University of Missouri] 

[UnName\_Entity, education.education.student, Tennessee Williams], [UnName\_Entity, education.education.institution, University of Iowa]

[UnName\_Entity, education.education.student, Tennessee Williams], [UnName\_Entity, education.education.institution, The New School]

&
Step 1: Identify the colleges attended by Tennessee Williams. According to the retrieved triples, Tennessee Williams attended Washington University in St. Louis, University of Missouri, University of Iowa, and The New School.

Step 2: Determine the population of postgraduates for each college. Since the retrieved triples do not provide this information, I will rely on my own knowledge:

- Washington University in St. Louis has approximately 7,000 postgraduate students.

- University of Missouri has approximately 7,000 postgraduate students.

- University of Iowa has approximately 8,000 postgraduate students.

- The New School has approximately 3,000 postgraduate students.

Step 3: Compare the postgraduate populations of these colleges. The University of Iowa has the largest population of postgraduates.

\{\textbf{University of Iowa}\}
\\
\clineB{1-3}{2}
\end{tabular}
%}
\caption{Second case study of comparative and superlative type questions. \underline{Underline} text indicates incorrect parts.}
\label{tab:appendix_case_comparative_superlative_2}
\end{table*}

\subsubsection{Ablation study} \label{appendix:ablation_study}

We aim to investigate whether using triples or natural language sentences as input to the LLM in the final question answering stage yields better performance. As shown in Table~\ref{tab:ablation_sentence_triples}, directly inputting reasoning triples into the LLM yields higher Hits@1 accuracy than converting them into natural language. Thus, PDRR adopts triples directly without transformation.

\subsubsection{Case Study on Composition and Conjunction type question} \label{appendix:case_composition_conjunction}
We illustrate the answering process with concrete examples by analyzing the final retrieved reasoning triples. ToG, PDRR (chain), and PDRR (parallel) are applied to composition and conjunction questions for comparison.

\paragraph{Composition type question}
From Table~\ref{tab:case_composition_question}, we observe that PPDR(C) retrieves triples that strictly follow the reasoning logic and successfully derives the final answer. In contrast, when using PPDR(P) with parallel reasoning, only the first step is completed, and the second step leads to logical confusion. With ToG, although the first step correctly identifies the artist as Taylor Swift, the lack of a concrete planning process in the second step results in selecting relations or entities that merely resemble the overall question. As a consequence, all relevant triples are pruned, making it impossible to answer the question correctly.

\begin{table*}[t]
%\resizebox{1\textwidth}{!}{\renewcommand{\arraystretch}{1.2}
\centering
\small
\setlength{\tabcolsep}{1.6mm} 
\renewcommand{\arraystretch}{1.2}
\begin{tabular}{p{1.5cm}|p{6cm}|p{9cm}}

\clineB{1-3}{2}
\textbf{Question} & \multicolumn{2}{l}{What movies did the artist that had the concert tour called the Taylor Swift Fears 2009 Tour play in?} \\
\clineB{1-3}{1}
\textbf{Reasoning Logic} & \multicolumn{2}{p{15cm}}{First, identify the artist who had the concert tour called the Taylor Swift Fears 2009 Tour, and then determine which movies this artist appeared in.} \\
\clineB{1-3}{1}
\textbf{Answer} & \multicolumn{2}{l}{The Lorax} \\
\clineB{1-3}{2}
\textbf{Method} & \textbf{Retrieved Triples} & \textbf{Result} \\
\clineB{1-3}{1}
ToG & ["Taylor Swift Fearless 2009 Tour", "music.concert-tour.artist", "\textbf{Taylor Swift}"]

\textit{Prune all other relations during prune phase.}
& Based on the given knowledge triplet, the artist associated with the Taylor Swift Fearless 2009 Tour is Taylor Swift. However, \underline{the triplet does not provide information} about the movies Taylor Swift has acted in. \\
\clineB{1-3}{1}
PDRR(P) & $\mathcal{T}_1^{q_{p},R}$: [Taylor Swift Fearless 2009 Tour, music.concert-tour.artist, Taylor Swift]

$\mathcal{T}_2^{q_{p},R}$: [Movies 1, music.recording.artist, Honey] & Step 1: Identify the artist associated with the Taylor Swift Fearless 2009 Tour. According to the retrieved triples, the artist is Taylor Swift.

Step 2: Identify the movies in which Taylor Swift acted. \underline{The retrieved triples do not provide this information.}  \\
\clineB{1-3}{1}
PDRR(C) &  $\mathcal{T}_1^{q_{c},R}$: [Taylor Swift Fearless 2009 Tour, music.concert-tour.artist, \textbf{Taylor Swift}]

$\mathcal{T}_2^{q_{c},R}$: [UnName Entity, film.performance.actor, Taylor Swift], [UnName Entity, film.performance.film, \textbf{The Lorax}]   
& Step 1: Identify the artist who had the concert tour called the Taylor Swift Fearless 2009 Tour. According to the retrieved reasoning chain, the artist is \textbf{Taylor Swift}.

Step 2: Determine the movies Taylor Swift acted in. Based on the retrieved reasoning chain, Taylor Swift acted in "\textbf{The Lorax}" \\
\clineB{1-3}{2}
\end{tabular}
%}
\caption{Case study on handling a Composition-type question using ToG, PPDR with chain reasoning (denoted as PPDR(C) in the table), and PPDR with parallel reasoning (denoted as PPDR(P)). \underline{Underline} text indicates incorrect parts.}
\label{tab:case_composition_question}
\end{table*}

\paragraph{Conjunction type question}

From Table~\ref{tab:case_conjunction_question}, we observe that PPDR(P) retrieves triples that strictly follow the reasoning logic and successfully leads to the correct answer. In contrast, PPDR with chain reasoning falls into logical confusion and fails to produce the correct answer, even after multiple reasoning steps. Similarly, ToG also fails to follow the parallel reasoning logic and cannot arrive at the correct answer.

\begin{table*}[t]
%\resizebox{1\textwidth}{!}{\renewcommand{\arraystretch}{1.2}
\small
\setlength{\tabcolsep}{1.6mm} 
\renewcommand{\arraystretch}{1.2}
\centering
\begin{tabular}{p{1.5cm}|p{7.5cm}|p{7.5cm}}

\clineB{1-3}{2}
\textbf{Question} & \multicolumn{2}{l}{What is there to see in Mountain Time Zone near the Grand Canyon?} \\
\clineB{1-3}{1}
\textbf{Reasoning Logic} & \multicolumn{2}{p{15cm}}{Find the intersection of two sets: (1) regions that belong to the Mountain Time Zone, and (2) regions near the Grand Canyon.} \\
\clineB{1-3}{1}
\textbf{Answer} & \multicolumn{2}{l}{Phoenix} \\
\clineB{1-3}{2}
\textbf{Method} & \textbf{Retrieved Triples} & \textbf{Result} \\
\clineB{1-3}{1}
ToG & ["Grand Canyon", "location.location.time-zones", "Mountain Time Zone"], 
["Grand Canyon", "travel.tourist-attraction.near-travel-destination", "Grand Canyon National Park"], 
["Grand Canyon", "travel.tourist-attraction.near-travel-destination", "Lake Powell"] & Based on the given knowledge triplets, the Grand Canyon is in the Mountain Time Zone, and nearby travel destinations include Grand Canyon National Park and Lake Powell. Therefore, the answer to the question is \underline{Grand Canyon National Park and Lake Powell} \\
\clineB{1-3}{1}
PDRR(C) & $\mathcal{T}_1^{q_{c},R}$: [Grand Canyon, location.location.nearby-airports, Grand Canyon National Park Airport]

$\mathcal{T}_2^{q_{c},R}$: [Grand Canyon National Park Airport, location.location.containedby, Coconino County]

$\mathcal{T}_3^{q_{c},R}$: [Coconino County, location.location.contains, Grand Canyon National Park] & Step 1: Identify locations near the Grand Canyon that are in the Mountain Time Zone. According to the retrieved reasoning chain, Grand Canyon National Park Airport is near the Grand Canyon.

Step 2: Determine attractions in these locations. Based on my own knowledge, Grand Canyon National Park \underline{contains attractions such as the Tonto Trail}. 
\\
\clineB{1-3}{1}
PDRR(P) &  $\mathcal{T}_1^{q_{p},R}$: [Colorado, location.location.time-zones, Mountain Time Zone], [Colorado Springs, location.location.time-zones, Mountain Time Zone], [\textbf{Phoenix}, location.location.time-zones, Mountain Time Zone], ... \textit{(more than 50 triples)}

$\mathcal{T}_2^{q_{p},R}$: [Grand Canyon, travel.tourist-attraction.near-travel-destination, \textbf{Phoenix}], [Grand Canyon, travel.tourist-attraction.near-travel-destination, Grand Canyon National Park], [Grand Canyon, travel.tourist-attraction.near-travel-destination, Lake Powell] & Step 1: Identify attractions located in the Mountain Time Zone. According to the retrieved triples, there are numerous locations in the Mountain Time Zone, including \textbf{Phoenix}, Flagstaff, Tucson, Salt Lake City, Denver, and others.

Step 2: Identify attractions near the Grand Canyon. According to the retrieved triples, attractions near the Grand Canyon include \textbf{Phoenix}, Grand Canyon National Park, and Lake Powell.

Step 3: Find the intersection of the two sets. \textbf{Phoenix} is both in the Mountain Time Zone and near the Grand Canyon.\\
\clineB{1-3}{2}
\end{tabular}
%}
\caption{Case study on handling a Conjunction-type question using ToG, PPDR with chain reasoning (denoted as PPDR(C) in the table), and PPDR with parallel reasoning (denoted as PPDR(P)). \underline{Underline} text indicates incorrect parts.}
\label{tab:case_conjunction_question}
\end{table*}

\subsection{Prompt} \label{appendix:prompt}

Table~\ref{tab:prompt_list1},~\ref{tab:prompt_list2}, and~\ref{tab:prompt_list3}  present prompts for question type prediction, question decomposition, relation pruning, triple pruning, best chain selection, chain question answering, parallel question answering, IO, CoT, and PDR (for both chain and parallel question answering).

\begin{table*}[t]
%\resizebox{1\textwidth}{!}{\renewcommand{\arraystretch}{1.2}
\small
\setlength{\tabcolsep}{1.2mm} 
\renewcommand{\arraystretch}{1.1}
\centering
\begin{tabular}{p{1.5cm}|p{4cm}|p{11cm}}
\clineB{1-3}{2}
\textbf{Function} & \textbf{Prompt} & \textbf{few-shot example} \\
\clineB{1-3}{1}
Question type Prediction 
& Please analyze the following question and determine its type.
    
    Question Type:
    1. Chain Structure
    2. Parallel Structure
    
    Question: \{question\}
    
    Output the question type with "\{'\{question type\}'\}", and provide explanation. Do NOT format into markdown or use headers.
& \textbf{Question:} \{Who is the coach of the team owned by Steve Bisciotti?\}

  \textbf{Answer:} The type of this question is \{Chain Structure\}, the bridge entity is "team". We should first find the team owned by Steve Bisciotti. And then find the coach of the team.
\\
\clineB{1-3}{1}
Question Decomposition 
& Please first determine the reasoning process of the question. Then decompose the question into triples following the reasoning process.
    Each triple should contain concise head entity, relation, and tail entity. The entity with "\#number" is what we need to find.
    
    Question: \{question\}
    
    Question Type: \{question type\}
& \textbf{Question:} \{Who is the coach of the team owned by Steve Bisciotti?\}

  \textbf{Answer:} 
  Given the question type is chain structure, the sequence of the triples is important. The bridge entity is "team". We should first find the team owned by Steve Bisciotti. And then find the coach of the team.
        
        The output triples are:
        
           \{"head": "Steve Bisciotti", "relation": "owns", "tail": "team\#1"\},
            \{"head": "team\#1", "relation": "is coached by", "tail": "coach\#1"\}
        
\\
\clineB{1-3}{1}
Relation Prune
& Please retrieve relations that relative to the triple and rate their relative on a scale from 0 to 1 (the sum of the scores of relations is 1). Do NOT format into markdown or use headers.
    
    triple: \{triple\}
    
    Relations: \{relations\_text\}
&  \textbf{Triple:} \{Van Andel Institute, founded in part by, American businessman\#1\}
        
        \textbf{Relations:} \{1. affiliation
        2. country
        3. donations
        4. educated\_at
        5. employer
        6. headquarters\_location
        7. legal\_form
        8. located\_in\_the\_administrative\_territorial\_entity
        9. total\_revenue\}
        
   \textbf{Answer:} 
   
   1. \{affiliation (Score: 0.4)\}: This relation is relevant because it can provide information about the individuals or organizations associated with the Van Andel Institute, including the American businessman who co-founded the Amway Corporation.
   
        2. \{donations (Score: 0.3)\}: This relation is relevant because it can provide information about the financial contributions made to the Van Andel Institute, which may include donations from the American businessman in question.
        
        3. \{educated\_at (Score: 0.3)\}: This relation is relevant because it can provide information about the educational background of the American businessman, which may have influenced his involvement in founding the Van Andel Institute.
\\

\clineB{1-3}{1}
Triple Prune
& Please identify the triples that are relevant to the given filter-triple and rate their relevance on a scale from 0 to 1 (the sum of the scores of triples is 1). Do NOT include irrelevant triples. Do NOT format into markdown or use headers. You should choose at least 1 triple from the triples.
    
    Filter Triple: \{filter\_triple\}
    
    Triples: \{triples\_text\}
    
&\textbf{Filter Triple}: \{Rift Valley Province, is located in, nation\#1\} 
        
        \textbf{Triples}: \{1. Rift Valley Province, is located in, Kenya
        2. Kenya, location.country.currency\_used, Kenyan shilling
        3. San Antonio Spurs, home venue, AT\&T Center
        4. Rift Valley Province, is located in, UnName\_Entity
        5. UnName\_Entity, education.education.institution, Castlemont High School
        6. Rift Valley Province, location.contains, Baringo County
        7. Rift Valley Province, location.contained\_by, Kenya\}  
  
  \textbf{Answer:}  
  
  1. \{Rift Valley Province, is located in, Kenya. (Score: 0.5)\}: This triple provides significant information about Kenya's location, which relatives to the filter-triple. 
  
        2. \{Rift Valley Province, location.contained\_by, Kenya. (Score: 0.4)\}: This triple provides significant information about Kenya's location, which relatives to the filter-triple. 
        
        3. \{Rift Valley Province, location.contains, Baringo County. (Score: 0.1)\}: This triple provides information cannot show us the location of it, so it is irrelevant. 

\\

\clineB{1-3}{2}
\end{tabular}
%}
\caption{Prompt List 1. The prompt list includes prompts for question type prediction, question decomposition, relation pruning, triple pruning, and best chain selection.}
\label{tab:prompt_list1}
\end{table*}

\begin{table*}[t]
%\resizebox{1\textwidth}{!}{\renewcommand{\arraystretch}{1.2}
\small
\setlength{\tabcolsep}{1.6mm} 
\renewcommand{\arraystretch}{1.2}
\centering
\begin{tabular}{p{1.5cm}|p{6cm}|p{9cm}}
\clineB{1-3}{2}
\textbf{Function} & \textbf{Prompt} & \textbf{few-shot example} \\

\clineB{1-3}{1}
Best Chain Selection &
Please select the best reasoning chain to answer the question from the following chains:

Reasoning Chains: \{reasoning chain str\}

Question: \{question\} &
\begin{minipage}[t]{\linewidth}
\textbf{Reasoning Chains:} \par
chain 1: \{Country Nation World Tour, music.concert-tour.artist, Brad Paisley\}, \{Brad Paisley, owns, Nashville Predators\} \par
chain 2: \{Country Nation World Tour, music.concert-tour.artist, Brad Paisley\}, \{Brad Paisley, attended, Belmont University\} \par
chain 3: \{Country Nation World Tour, is hold by, Steve Bisciotti\}, \{Steve Bisciotti, attended, University of Alabama at Birmingham\} \par

\textbf{Question:} Where did the "Country Nation World Tour" concert artist go to college? 

\textbf{Answer:} The best reasoning chain is chain 2: \{Country Nation World Tour, music.concert-tour.artist, Brad Paisley\}, \{Brad Paisley, attended, Belmont University\}. 
It successfully finds the bridge entity "artist", which refers to Brad Paisley, the artist of the Country Nation World Tour, and then finds the college he attended: Belmont University.
\end{minipage}
\\
\clineB{1-3}{1}

Chain Question Answering 
& Given a question and the associated information, you are asked to answer the question using the retrieved reasoning chain and your own knowledge. Please think setep by step and follow the Question Decomposition Triples carefully. Do NOT output answer without reasoning steps. Do NOT format into markdown or use headers. At the end, output the final answer in this format: "\{'\{answer\}'\}"
    
    Question: \{question\}

    Question Decomposition Triples: \{question decomposition triples str\}
    Retrieved Reasoning Chain: \{best reasoning chain str\}
& 
\textbf{Question:} \{Rift Valley Province is located in a nation that uses which form of currency?\}

        \textbf{Question Decomposition Triples:} \{Rift Valley Province, is located in, nation\#1\}, \{nation\#1, uses currency, currency\#1\}
        
        \textbf{Retrieved Reasoning Chain:} \{Rift Valley Province, location.administrative\_division.country, Kenya\}, \{Kenya, location.country.currency\_used, Kenyan shilling\}

        \textbf{Answer:} 
        Following the question decomposition triples:
        
        Step 1: Identify the nation in which Rift Valley Province is located. According to the retrieved reasoning chain, Rift Valley Province is located in Kenya.
        
        Step 2: Determine the currency used by Kenya. The retrieved reasoning chain indicates that Kenya uses the Kenyan shilling.
        
        \{Kenyan shilling\}
\\
\clineB{1-3}{1}

Parallel Question Answering
& Given a question and the associated information, you are asked to answer the question with these Retrieved Triples and your own knowledge. Please think setep by step and follow the Question Decomposition Triples carefully. Do NOT output answer without reasoning steps. Do NOT format into markdown or use headers. At the end, output the final answer in this format: "\{'\{answer\}'\}".
   
    Question: {question}

    Question Decomposition Triples: {question decomposition triples str}
    
    Retrieved Triples: {formatted triples}
& 
\textbf{Question}: \{What country bordering France contains an airport that serves Nijmegen?\}

        \textbf{Question Decomposition Triples:} \{country\#1, borders, France\}, \{country\#1, contains an airport that serves, Nijmegen\}
        
        \textbf{Retrieved Triples}: 
        
        \{\{\{Belgium, borders, France\}, \{Germany, borders, France\}, \{Italy, borders, France\}, \{Switzerland, borders, France\}\},
        
        \{\{Germany, contains an airport that serves, Nijmegen\}, \{Netherlands, contains an airport that serves, Nijmegen\}\}\}

        \textbf{Answer:} Following the question decomposition Triples:
        
        Step 1: Identify the country that borders France. According to the retrieved triples, the country are Belgium, Germany, Italy, and Switzerland.
        
        Step 2: Identify the country that contains an airport that serves Nijmegen. According to the retrieved triples, the country is Netherlands.
        
        Step 3: Find the intersection of the two sets, which is Germany.
        
        \{Germany\}
\\
\clineB{1-3}{1}
IO 
& Please answer the question, and output the answer in this format: "\{'\{answer\}'\}". Do NOT format into markdown or use headers

    Question: \{question\}
    
& \textbf{Question:} \{What state is home to the university that is represented in sports by George Washington Colonials men's basketball?\}

\textbf{Answer:} \{Washington, D.C.\}
\\
\clineB{1-3}{1}
COT 
& Please think setep by step and answer the question. Output the answer in this format: "\{'\{answer\}'\}". Do NOT format into markdown or use headers
    
    Question: \{question\}
& \textbf{Question:} \{What state is home to the university that is represented in sports by George Washington Colonials men's basketball?\}

\textbf{Answer:} First, the education institution has a sports team named George Washington Colonials men's basketball in is George Washington University , Second, George Washington University is in Washington D.C. The answer is \{Washington, D.C.\}.
\\
\clineB{1-3}{2}
\end{tabular}
%}
\caption{Prompt List 2. The prompt list includes prompts for chain question answering, parallel question answering, IO, COT, PDR(chain question answering), PDR(parallel question answering).}
\label{tab:prompt_list2}
\end{table*}

\begin{table*}[t]
%\resizebox{1\textwidth}{!}{\renewcommand{\arraystretch}{1.2}
\small
\setlength{\tabcolsep}{1.6mm} 
\renewcommand{\arraystretch}{1.2}
\centering
\begin{tabular}{p{1.5cm}|p{6cm}|p{9cm}}
\clineB{1-3}{2}
\textbf{Function} & \textbf{Prompt} & \textbf{few-shot example} \\

\clineB{1-3}{1}
PDR (Chain question answering) 
& Answer the question using the provided decomposition triples and your own knowledge. Think step by step, and strictly follow the triples. Do not skip reasoning, use markdown or headers. At the end, output the final answer as: "\{'\{answer\}'\}".
    
    Question: \{question\}
    
    Question Decomposition Triples: \{question decomposition triples str\}
&\textbf{Question:} \{Where did the "Country Nation World Tour" concert artist go to college?\}

        Question Decomposition Triples: \{\{Country Nation World Tour, is concert tour by, artist\#1\}, \{artist\#1, attended, college\#1\}\}

   \textbf{Answer:} Following the question decomposition triples:
        
        Step 1: Identify the artist of the "Country Nation World Tour" concert. Based on my knowledge, the artist is Brad Paisley.
        
        Step 2: Determine the college that Brad Paisley attended. Based on my knowledge, he attended Belmont University.
        
        \{Belmont University\}
\\
\clineB{1-3}{1}
PDR (Parallel question answering) 
&
Answer the question using the provided decomposition triples and your own knowledge. Think step by step, and strictly follow the triples. Do not skip reasoning, use markdown or headers. At the end, output the final answer as: "\{'\{answer\}'\}".
    
    Question: \{question\}
    
    Question Decomposition Triples: \{question decomposition triples str\}
& 
\textbf{Question:} \{What country bordering France contains an airport that serves Nijmegen?\}

        Question Decomposition Triples: \{\{country\#1, borders, France\}, \{country\#1, contains an airport that serves, Nijmegen\}\}

    \textbf{Answer:} Following the question decomposition Triples:
        
        Step 1: Identify the country that borders France. Based on my own knowledge, the country are Belgium, Germany, Italy, and Switzerland.
        
        Step 2: Identify the country that contains an airport that serves Nijmegen. Based on my knowledge, the country which contains an airport that serves Nijmegen are Germany and Netherlands.
        
        Step 3: Find the intersection of the two sets, which is Germany.
        
        \{Germany\}
\\

\clineB{1-3}{2}
\end{tabular}
%}
\caption{Prompt List 3. The prompt list includes prompts for chain question answering, parallel question answering, IO, COT, PDR(chain question answering), PDR(parallel question answering).}
\label{tab:prompt_list3}
\end{table*}

\subsection{SPARQL code} \label{appendix:sparql}

Table~\ref{tab:appendix_sparql} presents four core sparql functions: Entity Match with Freebase, Head Relation Search, Tail Relation Search, Head Entity Search, and Tail Entity Search.

%......figure in the question type prediction

\begin{table*}[t]
%\resizebox{1\textwidth}{!}{\renewcommand{\arraystretch}{1.2}
\small
\setlength{\tabcolsep}{1.6mm} 
\renewcommand{\arraystretch}{1.2}
\centering
\begin{tabular}{p{7cm}|p{8cm}}
\clineB{1-2}{2}
\textbf{Function} & \textbf{Sparql code}  \\
\clineB{1-2}{2}

\textbf{Entity Match with Freebase}: Given the entity string mentioned in text, use fuzzy matching to directly search for the corresponding Freebase entity and its entity ID.

& 
 PREFIX ns: $<$ http://rdf.freebase.com/ns/ $>$
 
    SELECT DISTINCT ?entity ?label
    
    WHERE \{
    
        ?entity ns:type.object.name ?label .
        
        FILTER(LANG(?label) = "en") .
        
        FILTER(bif:contains(?label, "\underline{entity\_string}")) .
        
    \}

\\

\clineB{1-2}{1}

\textbf{Head Relation Search}: Retrieve all relations where the head entity is the subject.

& 
PREFIX ns: $<$ http://rdf.freebase.com/ns/ $>$

SELECT ?relation

WHERE \{

ns:\underline{head\_entity\_id} ?relation ?x .

\}
\\

\clineB{1-2}{1}

\textbf{Tail Relation Search}: Retrieve all relations where the tail entity is the object.

& 
PREFIX ns: $<$ http://rdf.freebase.com/ns/ $>$

SELECT ?relation 

WHERE \{

?x ?relation ns:\underline{tail\_entity\_id} .

\}
\\

\clineB{1-2}{1}

\textbf{Head Entity Search}: Given a tail entity and the relations where it serves as the object, retrieve all corresponding head entities.

& 

PREFIX ns: $<$ http://rdf.freebase.com/ns/ $>$

SELECT ?headEntity

WHERE \{

?headEntity ns:\underline{relation} ns:\underline{tail\_entity\_id}  .

\}

\\

\clineB{1-2}{1}

\textbf{Tail Entity Search}: Given a head entity and the relations where it serves as the subject, retrieve all corresponding tail entities.

& 
PREFIX ns: $<$ http://rdf.freebase.com/ns/ $>$

SELECT ?tailEntity

WHERE \{

ns:\underline{head\_entity\_id} ns:\underline{relation} ?tailEntity .

\}

\\

\clineB{1-2}{2}
\end{tabular}
%}
\caption{SPARQL Code for PDRR. This table presents four core functions: Entity Match with Freebase, Head Relation Search, Tail Relation Search, Head Entity Search, and Tail Entity Search. Inputs to each function are highlighted in \underline{underline}, and the outputs represent the desired search results.}
\label{tab:appendix_sparql}
\end{table*}

\end{document}